\documentclass[sigconf]{acmart}

\usepackage{booktabs} 

\usepackage{algorithm}
\usepackage{algorithmic}

\usepackage{fancyhdr,amsmath,amsthm,amssymb,bm} 
\usepackage{graphicx,caption,verbatim}
\usepackage{bm,bbm,float,mwe,dsfont,color,url,diagbox,multirow}
\usepackage{enumitem}

\def \bal#1\eal{\begin{align}#1\end{align}}
\def \bals#1\eals{\begin{align*}#1\end{align*}}

\def\Loss{\mathcal{L}}

\def\real{\mathbb{R}}

\def\F{\textnormal{F}}

\let\hat\widehat
\let\tilde\widetilde

\def\Loss{\mathcal{L}}
\newcommand{\minimize}{\operatornamewithlimits{minimize}}
\newcommand{\argmin}{\operatornamewithlimits{argmin}}

\graphicspath{{fig/}}

\setcopyright{none}

\begin{document}
\title[Multitask Learning using Task Clustering]{Multitask Learning using Task Clustering with Applications to Predictive Modeling and GWAS of Plant Varieties}

\author{Ming Yu}
\affiliation{%
  \institution{Booth School of Business,\\ The University of Chicago}
  \city{Chicago} 
  \state{IL}
  \country{USA} 
}
\email{ming93@uchicago.edu}

\author{Addie M. Thompson}
\affiliation{%
  \institution{Dept. of Agronomy,\\ Purdue University}
  \city{W.  Lafayette} 
  \state{IN}
  \country{USA}
}
\email{thomp464@purdue.edu}

\author{Karthikeyan Natesan Ramamurthy}
\affiliation{%
  \institution{IBM T.J. Watson Research Center}
  \city{Yorktown Heights} 
  \state{NY}
  \country{USA}}
\email{knatesa@us.ibm.com}

\author{Eunho Yang} 
\affiliation{%
 \institution{KAIST}
 \city{Daejeon} 
 \country{South Korea}}
\email{eunhoy@kaist.ac.kr}

\author{Aurélie C. Lozano}
\affiliation{%
  \institution{IBM T.J. Watson Research Center}
  \city{Yorktown Heights} 
  \state{NY}
  \country{USA}}
\email{aclozano@us.ibm.com}

\renewcommand{\shortauthors}{M. Yu et al.}

\begin{abstract}
Inferring predictive maps between multiple input and multiple output variables or tasks has innumerable applications in data science. Multi-task learning attempts to learn the maps to several output tasks simultaneously with information sharing between them. We propose a novel multi-task learning framework for sparse linear regression, where a full task hierarchy is automatically inferred from the data, with the assumption that the task parameters follow a hierarchical tree structure. The leaves of the tree are the parameters for individual tasks, and the root is the global model that approximates all the tasks. We apply the proposed approach to develop and evaluate: (a) predictive models of plant traits using large-scale and automated remote sensing data, and (b) GWAS methodologies mapping such derived phenotypes in lieu of hand-measured traits. We demonstrate the superior performance of our approach compared to other methods, as well as the usefulness of discovering hierarchical groupings between tasks. Our results suggest that richer genetic mapping can indeed be obtained from the remote sensing data. In addition, our discovered groupings reveal interesting insights from a plant science perspective.
\end{abstract}

%
%
%


\maketitle

%
%
%
%
%
%

\section{Introduction}
Several problems in data science necessitate discovery of a predictive mapping from input to output data, such that it can be generalized to new inputs. In many applications, there can be more than one output variable and they can be related to each other, and it is also desirable for the mappings to have some structure. Multi-output regression models generalize single-output regression models to learn predictive relationships between multiple input and multiple output variables, also referred to as tasks \cite{borchani2015survey}. Multi-task learning attempts to learn several the inference tasks simultaneously, and the assumption here is that an appropriate sharing of information  can benefit all the tasks \cite{caruana1998multitask, obozinski2006multi}. There are several ways to define task-relatedness; the parameters can be close to each other \cite{evgeniou2004regularized}, they can share a common prior \cite{daume2009bayesian}, or they can share a common latent feature representation  \cite{caruana1998multitask}. We will only focus on multi-task learning under the linear sparse modeling framework.

The implicit assumption that sharing all features for all tasks can be excessive and can ignore the underlying specificity of the mappings. There have been several extensions to multi-task learning that address the problem of how to, and with whom to share between the tasks \cite{jalali2010dirty, kang2011learning, kim2010tree, swirszcz2012multi, jacob2009clustered, kumar2012learning}. The authors in \cite{jalali2010dirty} propose a \emph{dirty} model for feature sharing among tasks, wherein a linear superposition of two sets of parameters - one that is common to all tasks, and one that is task-specific is used. A variant of this was proposed in \cite{swirszcz2012multi} with elementwise product between the two parameter sets. The approach proposed in \cite{kumar2012learning} learns to share by defining a set of \emph{basis task parameters} and posing the task-specific parameters as a sparse linear combination of these. In \cite{jacob2009clustered} and \cite{kang2011learning}, the authors assume that the tasks are clustered into groups and proceed to learn the group structure along with the task parameters using a convex and an integer quadratic program respectively. In contrast to the above approaches, \cite{kim2010tree} leverages a predefined tree structure among the output tasks (e.g. using hierarchical agglomerative clustering) and imposes group regularizations on the task parameters based on this tree.

In this work, we propose to simultaneously learn the task parameters of a multi-task linear regression model as well as the structured relationship between the tasks in a fully supervised manner. We assume that the task parameters follow a general hierarchical tree structure, and hence this approach learns the full task sharing hierarchy automatically. We adopt such a structure as it is natural to model task relatedness, similarly as in hierarchical clustering and it is also easily interpretable. Our framework is motivated by several applications such as in genome wide association analysis with multiple output phenotypes \cite{schifano2013genome}, and predictive modeling of plant traits with remote sensed data in automated phenotyping \cite{ramamurthy2016predictive} to name a few. To the best of our knowledge, ours is the first attempt in the literature to learn a hierarchical task clustering structure in a supervised manner. Our proposed approach constrains the parameters of each task to be sparse and incorporates an additional regularization on the task parameters inspired by convex clustering \cite{hocking2011clusterpath}. Similar to convex clustering, a continuous regularization path of solutions can be obtained and each point in the path corresponds to a threshold for cutting the task parameter tree.

The main contributions of this paper are as follows. We propose a structured multi-task learning estimator that can learn a tree structure among the tasks while simultaneously estimating the task parameters. We present a proximal decomposition approach to efficiently solve the resulting convex problem, and show its numerical convergence. We provide statistical guarantees and study the asymptotic properties of our estimator. Our approach is validated empirically on simulated data. We also explore the applications of the proposed method in two high-throughput phenotyping problems: (a) predictive modeling of plant traits using large-scale and automated remote sensing data, and (b) Genome-Wide Association Studies (GWAS) mapping such derived phenotypes in lieu of hand-measured traits. Initial results are promising and show that our method produces high predictive accuracy while revealing groupings that are insightful.

\section{Background}


\paragraph{Multi-task Learning.}
Consider the multi-task linear regressions: for each task $s=1, \hdots, k$
\[ y = x^T \theta_s^* + \epsilon,\]
for where $y \in \real$, $x \in \real^p$, and $\epsilon \sim \mathcal{N}(0,\sigma^2)$.  
Suppose we are given $n_s$ i.i.d. samples $\{x_i, y_i\}_{i=1}^{n_s}$ for task $s$. We will overload notation and collate these observations using vector notation as:
\[ y_s = X_s \theta^*_s + \epsilon_s,\]
where $y_s \in \real^{n_s}$, $X_s \in \real^{n_s \times p}$, and $\epsilon_s \in \real^{n_s}$. Throughout the paper, we concatenate $k$ parameters in columns to form $\Theta^* \in \real^{p\times k}$ (i.e. $s$-th column, $\Theta^*_s$, is $\theta^*_s$), and assume that all tasks have the same number of observations $n$ for notational simplicity. Note that in the multi-response model where the design matrix $X_s$ is shared across all tasks, we have $Y = X \Theta^* + E$ in a simpler form with $Y \in \real^{n\times k}$, $X\in \real^{n\times p}$ and $E \in \real^{n\times k}$.
Generally, multi-task learning solves the following optimization problem 
\begin{align}
\minimize_{\Theta}\sum_{s=1}^k \  \|y_s - X_s\Theta_s\|_2^2  +  \lambda_1 \| \Theta_s\|_1 + \Omega(\Theta)
\end{align}
where the penalty $\Omega(\Theta)$ on the $k$ task parameters encourages information sharing.

\paragraph{Convex Clustering.}
Clustering is a fundamental \emph{unsupervised} problem which is broadly used in many scientific applications. Given $k$ data points $x_1, \hdots, x_k \in \real^{p}$ (our choice of notation here is deliberate for the next section), \citet{Lindsten2011,Hocking2011} propose a convex optimization problem for clustering $k$ points:
\begin{align}\label{EqnCVXClustsering}
	\minimize_{v} \frac{1}{2} \sum_{s=1}^k \|x_s - v_s\|_2^2 + \lambda \sum_{s<t} w_{st} \| v_s - v_t\|_2	
\end{align}
where $\lambda$ is a positive tuning parameter, $w_{st}$ is a nonnegative weight. Note that for the second term, other penalties rather than $\ell_2$ norm have been also considered. As discussed in \citet{chi2015splitting}, each point constitutes an independent cluster when $\lambda$ is small enough, however, as $\lambda$ increases, the cluster centers start to combine.

\section{Multitask Learning with Task Clustering}

Our goal is to jointly infer (a) the linear regression parameter matrix $\Theta^*$ allowing for high-dimensional sampling settings where the number of observations is possibly much larger than the problem dimension $p$; (b) a general hierarchical tree structure among the tasks.

To accomplish this goal, we propose to solve the following regularized regression problem:
\begin{align}\label{EqnMethod}
\minimize_{\Theta}\sum_{s=1}^k \ & \Big\{ \|y_s - X_s\Theta_s\|_2^2  +  \lambda_1 \| \Theta_s\|_1\Big\} \nonumber\\
	& + \lambda_2 \sum_{s<t} w_{st} \| \Theta_s - \Theta_t \|_2 \, .
\end{align}



The above formulation has three terms. %
The first term is the squared loss. While we focus on linear regression, our approach can be readily generalized and employ other loss functions such as the logistic loss, the $\ell_1$ loss, etc.

The second term encourages the sparsity of the regression coefficient matrix $\Theta$. Clearly, this can be generalized to a more sophisticated sparsity structure, e.g., group sparsity, based on the needs of the application.

The third term encourages tasks to share the same value of their parameter vectors. It can be viewed as a generalization of the fused lasso penalty~\cite{Tibshirani05} wherein we \emph{fuse} the parameter vectors rather than the scalar coefficients. This is identical to the \emph{convex clustering} penalty~\cite{chi2015splitting} in \eqref{EqnCVXClustsering}, but it forms clusters for task parameters rather than data points. Hence,  brought into our framework, this penalty induces clustering of task parameters, effectively learning a hierarchical tree of these parameters.

$\lambda_1$ and $\lambda_2$ are tuning parameters that control the degree of individual task sparsity and task sharing respectively. As in \eqref{EqnCVXClustsering}, when $\lambda_2$ is small enough, each task parameter $\Theta_s$ is allowed to focus on minimizing individual loss function $\|y_s - X_s\Theta_s\|_2^2  +  \lambda_1 \| \Theta_s\|_1$. However, as $\lambda_2$ increases, the relative importance of third term increases and task parameters start to combine. In this sense, our formulation \eqref{EqnMethod} can be viewed as \emph{supervised} clustering problem. Setting these parameters is essential to decide the behavior of \eqref{EqnMethod}, and they can be tuned e.g. via cross validation. Finally, $w_{st}$ are optional non-negative weights that can be imposed to reflect prior knowledge on the degree of relatedness between each pair of tasks. 

The choice of the weights can dramatically affect the complexity of the minimization problem. Typically it is desirable to have sparse weights $w_{st}$: only a small portion of them are non-zero. As suggested in \cite{chi2015splitting}, we can set $w_{st} = 1^\kappa_{st} \cdot \exp(-\phi\| \bm{y}_s - \bm{y}_t \|_2^2) $, where $1^\kappa_{st}$ is 1 if $t$ is among $s$'s $\kappa$-nearest-neighbors or vice versa and 0 otherwise. The constant $\phi$ is nonnegative; $\phi = 0$ corresponds to uniform weights for all $\kappa$ neighbors.

\subsection{Optimization via Proximal Decomposition}

In this section we describe an optimization algorithm to solve our formulation \eqref{EqnMethod}. Since the task loss function for linear regression problem is convex, our formulation overall is also convex in $\Theta$, so it can be solved using modern convex optimization approaches. The story here can be trivially generalized to general loss function $\ell(\Theta)$ beyond squared loss, as long as it is convex.

Here we adopt the proximal decomposition method introduced in \cite{combettes2008proximal}. This is an efficient algorithm for minimizing the summation of several convex functions.  Our objective function involves
%
 $m = 2 + \# \{(s,t):w_{st} > 0\}$ such functions (Note that $w_{st}$ is usually sparse so we do not have to deal with $\mathcal{O}(k^2)$ terms):
  \begin{align*}
&f_1(\Theta)=\sum_{s=1}^k \|y_s - X_s \Theta_s \|_2^2,\\
&f_2(\Theta)= \lambda_1  \sum_{s=1}^k \| \Theta_s\|_1,\\
&f_{s,t}(\Theta) =  \lambda_2 \sum_{s<t} w_{st} \| \Theta_s - \Theta_t \|_2 : \{(s,t):w_{s,t}>0\}.\\
\end{align*}
At a high level, the algorithm iteratively applies proximal updates with respect to the above functions. We now describe the details of the algorithm as summarized in Algorithm~\ref{proximal}. 
 We stack the regression matrix $\Theta$ into a column vector $(\Theta_1; ...; \Theta_k) \in \mathbb{R}^{pk}$.
%
%
In the algorithm $l$ denote the iteration number. Each $\tilde\Theta_{1,l}$,  $\tilde\Theta_{2,l}$ or  $\tilde\Theta_{st,l}$ is a $pk$ dimensional column vector, corresponding to our parameter $(\Theta_1; \ldots; \Theta_k)$. The procedure has two additional parameters $\gamma$ and $\mu_l.$ In practice we can set $\mu_l=1$ for each iteration $l.$
%
%
`prox' denotes the proximal operator:
$
\text{prox}_f  b  = \mathop{\text{argmin}}\limits_a \Big( f(a) + \frac 12 \|b-a\|^2 \Big),
$ where $a$ and $b$ are vectors.


We iteratively apply the proximal updates for $f_1,f_2$ and $f_{s,t}$ until convergence. The specific update rules are as follows.

\begin{itemize}
\item  Update for $f_1$: 
Let $(a_1; ...; a_k) = \text{prox}_{\sigma f_1}(b_1; ...; b_k),$
For each $s=1,\ldots k$, we have $$a_s = (\sigma X_s^TX_s + \frac 12 I_p)^{-1} \cdot (\sigma X_s^Ty_s + \frac 12 b_s).$$
This step corresponds to the closed-form formula of a ridge regression problem. For very large $p$ we can employ efficient approaches such as \cite{mcwilliams2014loco} and \cite{lu2014fast}.

\item Update for $f_2$: 
Let $(a_1; ...; a_k) = \text{prox}_{\sigma f_2}(b_1; ...; b_k),$
For each $s \in \{1,\ldots,k\}$, $j\in \{1,\ldots,p\}$,
$$[a_{s}]_j = \Bigg[ 1-\frac{\lambda_1\sigma}{ | [b_{s}]_j|} \Bigg]_{+} \cdot [b_{s}]_j.$$

\item Updates for $f_{s,t}$: 
Let $(a_1; ...; a_k) = \text{prox}_{\sigma f_{st}}(b_1; ...; b_k),$
For each $(s,t) \in \{1,\ldots, k \}^2$ with $w_{st} > 0$, let $c_{st} = \frac{\sigma\lambda_2}{\|b_s - b_t\|}$, 
\begin{equation*}
\begin{aligned}
a_s &= (1-c_{st})\cdot b_s + c_{st} \cdot b_t, \\
a_t &= (1-c_{st})\cdot b_t + c_{st} \cdot b_s.
\end{aligned}
\end{equation*}
\end{itemize}


\begin{algorithm}[tb]
   \caption{}
   \label{proximal}
\begin{algorithmic}
   \STATE {\bfseries Initialize:} 
   
   \quad $\gamma \in (0,\infty)$, $\tilde\Theta_{1,0} \in \mathbb{R}^{pk},  $ $\tilde\Theta_{2,0} \in \mathbb{R}^{pk} $\\
   \quad $\tilde\Theta_{st,0} \in \mathbb{R}^{pk},  \text{ for }(s,t) : w_{st}>0$
   \\
   \vspace{0.6mm}
   \quad $\hat\Theta_0 = \frac{1}{m} (\tilde\Theta_{1,0} +\tilde\Theta_{2,0})+ \frac{1}{m} \sum_{s,t : w_st>0} \tilde\Theta_{st,0}$\\
   \vspace{1.5mm}
   \FOR{$l=0, 1, 2, 3, ...$ }
   \STATE $p_{1,l} = \text{prox}_{\sigma f_1} \tilde\Theta_{1,l}$ where $\sigma = m \gamma $
    \STATE $p_{2,l} = \text{prox}_{\sigma f_2} \tilde\Theta_{2,l}$ where $\sigma = m \gamma $
   \FOR{$s,t \text{ s.t. } w_{st} >0$}
   \STATE $p_{st,l} = \text{prox}_{\sigma f_{st}} \tilde\Theta_{st,l}$ where $\sigma =m \gamma $
   \ENDFOR
   \STATE $p_l = \frac{1}{m}(p_{1,l}+p_{2,l}) + \frac{1}{m} \sum_{s,t:w_{st}>0} p_{st,l}$
   \STATE $\mu_l \in (0,2)$
   \STATE $\tilde\Theta_{1,l+1} = \tilde\Theta_{1,l} + \mu_l(2p_l - \hat\Theta_l - p_{1,l})$
    \STATE $\tilde\Theta_{2,l+1} = \tilde\Theta_{2,l} + \mu_l(2p_l - \hat\Theta_l - p_{2,l})$
    \FOR{$s,t \text{ s.t. } w_{st} >0$}
   \STATE $\tilde\Theta_{st,l+1} = \tilde\Theta_{st,l} + \mu_l(2p_l - \hat\Theta_l - p_{st,l})$
   \ENDFOR
   \STATE $\hat\Theta_{l+1} = \hat\Theta_l + \mu_l(p_l - \hat\Theta_l)$
   \ENDFOR
   \vspace{1.5mm}
   \STATE {\bfseries Reshape $\hat\Theta$ to get $\Theta$.} 
\end{algorithmic}
\end{algorithm}

\subsection{Numerical Convergence}
The following proposition characterizes the convergence of Algorithm~\ref{proximal}. It is a direct consequence of Theorem 3.4. in \cite{combettes2008proximal}. 
\begin{proposition} \label{prop:conv}
Let $G$ be the set of solutions to problem~\eqref{EqnMethod} and let $(\hat\Theta_l)_{l \in \mathbb{N}}$ be a sequence generated by Algorithm~\ref{proximal}. Then provided that $\sum_{l \in \mathbb{N}} \mu_l(2-\mu_l) = +\infty$
Then $G \neq \emptyset$ and $(\hat\Theta_l)_{l \in \mathbb N}$ converges weakly to a  point in $G.$
\end{proposition}
The proof is presented in the appendix.

\section{Theoretical Guarantees}

In this section, we provide the asymptotic properties of our estimators \eqref{EqnMethod} that can be understood as the extension of those for vanilla Lasso \citep{Knight00} and fused Lasso \citep{Tibshirani05}. Following the strategy in~\cite{Knight00,Tibshirani05}, we assume in the main statement that the feature dimension $p$ is fixed as the number of samples $n$ approaches infinity. Though restrictive, this setting is more effective to illustrate the dynamics of methods. 

Before providing the main statement, we first introduce some quantities for clear representation. Let $\Theta^*$ be the population minimizer of the risk:
$\Theta^* \in \argmin_{\Theta}\sum_{s=1}^k \ \{ {\bf E} \| Y - X_s \Theta\|_\F^2  +  \lambda_1 \| \Theta_s\|_1 \} + \lambda_2 \sum_{s<t} w_{st} \| \Theta_s - \Theta_t \|_2$. Given the population or target parameter $\Theta^*$, we define two functions corresponding to two types of regularizers. Specifically, let $F(U ; \Theta)$ be the limit definition of derivative for individual $\ell_1$ regularization terms: $\lim_{h\rightarrow +0} \frac{\sum_{s=1}^k\{ \|\Theta_s + h U_s \|_1 - \|\Theta_s\|_1\}}{h}$, where $U \in \real^{p\times k}$. Then, $F(U;\Theta^*)$ can be rewritten as         	 
\begin{align}\label{EqnVL1}
	F(U; \Theta^*) := \sum_{s=1}^k \sum_{j=1}^p \ & U_{js} \text{sgn}(\Theta^*_{js}) \, {\bf I}(\Theta^*_{js}\neq 0) \nonumber\\
	& + |U_{js}| \, {\bf I}(\Theta^*_{js} = 0)
\end{align}
where ${\bf I}(\cdot)$ is the indicator function, and $\Theta^*_{js}$ (and $U_{js}$ respectively) is the $j$-th feature weight of the target parameter $\Theta^*_s$. Similarly, let $G_{st}(U_s, U_t \, ; \, \Theta_s, \Theta_t)$ be the limit definition of derivative for a single fusion term $w_{st}\|\Theta_s-\Theta_t\|_2$, defined as
\[
\lim_{h\rightarrow +0}~\frac{w_{st}\|\Theta_s + h U_s - \Theta_t - h U_t\|_2 - w_{st}\|\Theta_s - \Theta_t\|_2}{h}.
\] Then, it can be easily verified that the sum of $G_{st}$ over all fusion terms is written as
\begin{align}\label{EqnVL2}
	G(U; \Theta^*):= \sum_{s<t} & w_{st}  \frac{(\Theta^*_s-\Theta^*_t)^\top(U_s-U_t)}{\|\Theta^*_s-\Theta^*_t\|_2} \, {\bf I}(\Theta^*_s \neq \Theta^*_t) \nonumber\\
	& + w_{st}\|U_s-U_t\|_2 \, {\bf I}(\Theta^*_s = \Theta^*_t) \, .
\end{align}
 Armed with these quantities, the following theorem characterizes the limiting distribution on optimal solution of \eqref{EqnMethod}, $\widehat{\Theta}$.
\begin{theorem}\label{ThmMain}
	Consider a sequence of $\lambda_1$ and $\lambda_2$ such that $\lambda_1/\sqrt{n} \rightarrow \lambda_1^0 \geq 0$ and $\lambda_2/\sqrt{n} \rightarrow \lambda_2^0 \geq 0$ respectively, as $n \rightarrow \infty$. Suppose that $C_s := \lim_{n\rightarrow \infty} \left(\frac{1}{n} X^\top_s X_s \right)$ for every task $s = 1,\hdots,k$, is non-singular. Then, the estimator $\widehat{\Theta}$ in \eqref{EqnMethod} satisfies 
	\begin{align}
		\sqrt{n}\left(\widehat{\Theta}-\Theta^*\right) \overset{d}{\rightarrow} \argmin_{U} V(W,U)
	\end{align}  
	where 
		$V(W,U) := \sum_{s=1}^k \{-2 U_s ^\top W_s + U_s^\top C_s U_s \} + \lambda_1^0 F(U; \Theta^*) + \lambda_2^0 G(U; \Theta^*)$ using definitions in \eqref{EqnVL1} and \eqref{EqnVL2}, 		
	and $W \in \real^{p\times k}$ has an $p\times k$ normal distribution whose s-th column follows $N(0,\sigma^2C_s)$.
\end{theorem}

Given the construction of \eqref{EqnVL1} and \eqref{EqnVL2}, the proof of Theorem \ref{ThmMain} follows easily from the line of previous works e.g. \cite{Knight00} or \cite{Tibshirani05}. 

The theorem confirms that the joint limiting distribution of $\sqrt{n}(\widehat{\Theta}-\Theta^*)$ will have probability concentrated on the line $U_s = U_t$ if $\Theta^*_s = \Theta^*_t$, $w_{st} >0$ and $\lambda_2^0>0$, in addition to a lasso-type effect (as in \cite{Knight00}) on univariate limiting distributions when $\lambda_1^0 >0$. 

Note that in case of a `multi-output' setting where the design matrix $X$ is shared across all tasks, Theorem \ref{ThmMain} holds with the simpler form of $V(W,U)$: $-2 \textnormal{Tr}(U^\top W) + \textnormal{Tr}(U^\top C U) + \lambda_1^0 F(U; \Theta^*) + \lambda_2^0 G(U; \Theta^*)$. 

Note also that the extension of Theorem \ref{ThmMain} beyond least squares 
\begin{align}
\minimize_{\Theta}\sum_{s=1}^k \ & \Big\{ \Loss(\Theta_s; y_s, X_s) +  \lambda_1 \| \Theta_s\|_1\Big\} \nonumber\\
	& + \lambda_2 \sum_{s<t} w_{st} \| \Theta_s - \Theta_t \|_2 \, , \nonumber
\end{align} where $\Loss$ is a general loss function (e.g. logistic loss)
is also trivially attainable following \cite{Rocha2009}.

\section{Simulated Data Experiments}

\begin{figure*}
\minipage{0.4\textwidth}
\begin{center}
\centerline{\includegraphics[width=\columnwidth]{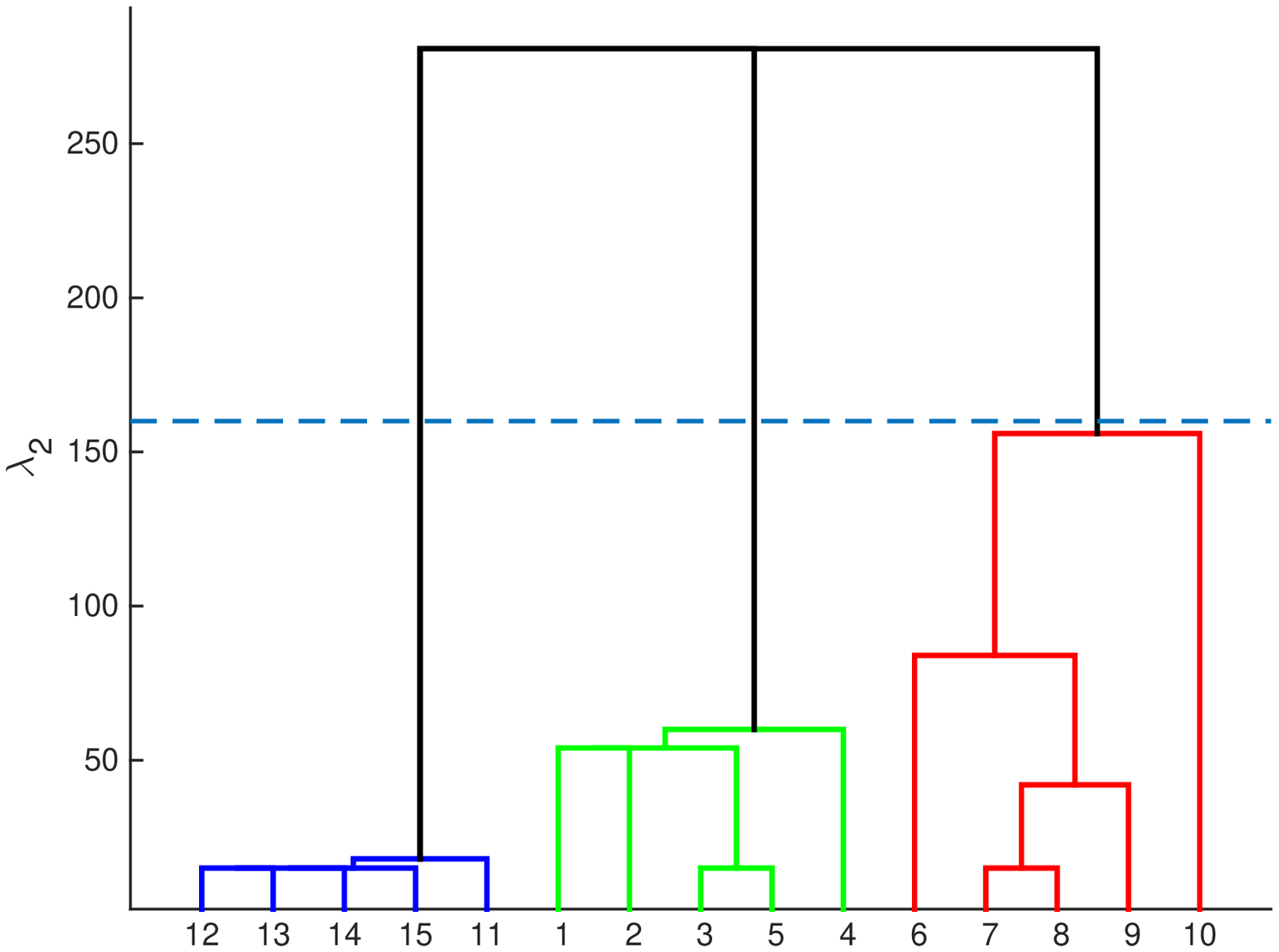}}
\caption{Tree structure learnt by our method}
\label{fig:example}
\end{center}
\endminipage\hfill
\minipage{0.4\textwidth}
\begin{center}
\centerline{\includegraphics[width=\columnwidth]{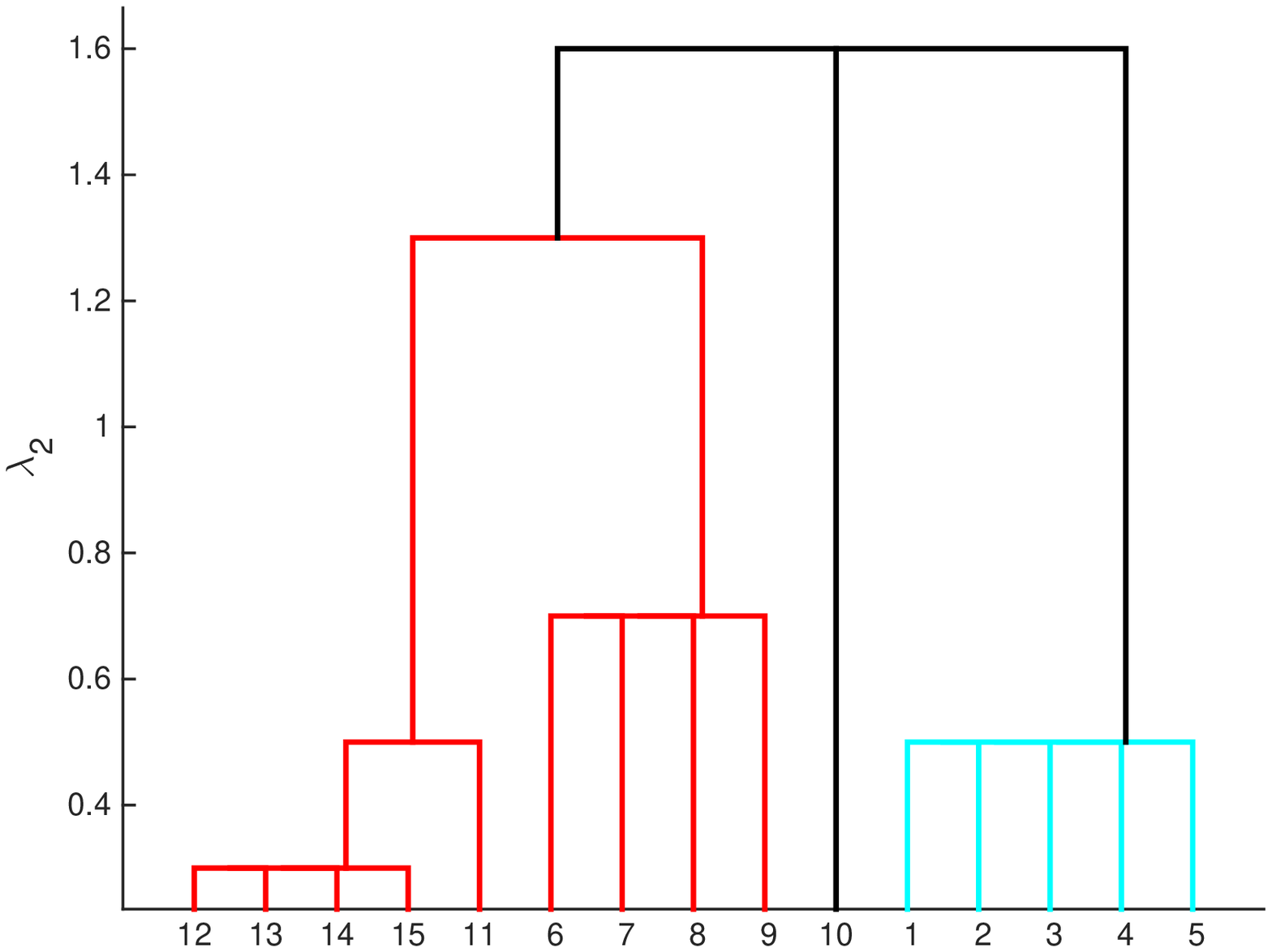}}
\caption{Tree structure learnt by post-clustering of Single Task Lasso}
\label{fig:examplesingle}
\end{center}
\endminipage\hfill

\end{figure*}

As a sanity check, we evaluate our approach on synthetic datasets and compare with the following baselines:

\begin{itemize}
\item {\bf Single task}: a baseline approach, where the tasks are learned separately via Lasso.
\item {\bf No group MTL}~\cite{obozinski2006multi}: the traditional multitask approach using group lasso penalty, where all tasks are learned jointly and the same features are selected across tasks.
\item {\bf Pre-group MTL}: Given the \emph{true} number of groups, first partition the tasks purely according to the correlation among responses and then apply \emph{No group MTL} in each cluster.
\item {\bf Kang et al}~\cite{kang2011learning}: Mixed integer program learning a shared feature representations among tasks, while simultaneously determining ``with whom'' each task should share. We used the code provided by the authors of~\cite{kang2011learning} and used the true number of tasks.
\item {\bf Tree-guided group Lasso}~\cite{kim2010tree}: Employs a structured penalty function induced from a predefined tree structure among responses, that encourages multiple correlated responses to share a similar set of covariates. We used the code provided by the authors of~\cite{kim2010tree} where the tree structure is obtained by running a hierarchical agglomerative clustering on the responses.
\end{itemize}

We consider $n=20$ samples, $p = 50$ features, and three groups of tasks. Within each group there are 5 tasks whose parameter vectors are sparse and identical to each other. We generate independent train, validation, and test sets. For each method, we select the regularization parameters using the validation sets, and report the root-mean-squared-error (RMSE)  of the resulting models on the test sets. We repeat this procedure 5 times. The results are reported in Table \ref{RMSE}.  From the table we can see that the \emph{Single task} method has the largest RMSE as it does not leverage task relatedness; \emph{No group MTL} has slightly better RMSE; both \emph{Kang et al.} and \emph{Tree guided group Lasso} get some improvement by considering structures among tasks; \emph{Pre-group MLT} achieves a better result, mainly because it is given the true number of groups and for these synthetic datasets it is quite easy to obtain good groups via response similarity, which might not necessarily be the case with real data and when the predictors differ from task to task. Our method achieves the smallest RMSE, outperforming all  approaches compared.
We also report the running times of each method, where we fix the tolerance parameters to $10^{-5}$ for comparison. Though the timing for each method could always be improved using more effective implementations, the timing result reflect the algorithmic simplicity of the steps in our approach compare to e.g. the mixed integer program of~\cite{kang2011learning}. 

\begin{table}[t]
\caption{RMSE for different comparison methods}
\label{RMSE}
\vskip 0.15in
\begin{center}
\begin{small}
\begin{tabular}{lccc}
\hline
Method & RMSE & std & time\\
\hline
Single task &  5.665 & 0.131 & 0.02\\
No group multitask learning &  5.520 & 0.115 & 0.05\\
Pre-group multitask learning &  5.256 & 0.117 & 0.10\\
Kang et al & 5.443 & 0.096 & $>10$\\
Tree guided group Lasso &  5.448 & 0.127 & 0.03\\
Ours &  \bf{4.828} & 0.117 & 0.16\\
\hline
\end{tabular}
\end{small}
\end{center}
\vskip -0.1in
\end{table}

An example of tree structure learnt by our approach is shown in Figure~\ref{fig:example}. It is contrasted with Figure~\ref{fig:examplesingle}, which depicts the structure obtained \emph{a-posteriori} by performing hierarchical clustering on the regression matrix learnt by \emph{Single Task.} The simulation scenario is the same as before except that the non-zero coefficients in each true group are not equal but sampled as $0.5 + N(0,1)/3,$ where $N$ denote the normal distribution.  As can be seen from Figure~\ref{fig:example}, no matter what $\lambda_2$ is, our approach never makes a mistake in the sense that it never puts tasks from different true groups in the same cluster. For $\lambda_2>150$ our approach recognizes the true groups. As $\lambda_2$ becomes very large there are no intermediary situation where two tasks are merged first. Instead all tasks are put in the same cluster.  We see tasks $\{3,5\}$ merge first in group $\{1-5\}$ and $\{7,8,9\}$ merge first in group $\{6-10\}.$ This corresponds to the fact that tasks $\{3,5\}$ have largest correlation among group $\{1-5\}$ and $\{7,8,9\}$ has largest correlation among group $\{6-10\}.$
We can see in Figure~\ref{fig:examplesingle} that for \emph{Single Task} post clustering, task 10 does not merge with $\{6-9\}.$


%

\paragraph{Impact of the weights $w_{st}.$}
One might argue that our approach relies on ``good'' weights $w_{st}$ among tasks. However, it turns out that it is quite robust to the weights. 
Recall that we select the weight $w_{st}$ by $\kappa$-nearest-neighbors. In this synthetic dataset, we have 5 tasks in each group so the most natural way is to set $\kappa = 4$. We also try setting $\kappa = 2,3,5,6$ and see how this affects the result. The test RMSEs for different $\kappa$'s are given in Table \ref{kappa}.
From the table we see that although the best performance is when we select $\kappa = 4$, our method is quite robust to the choice of weights, especially when $\kappa$ is smaller than the natural one. 
When $\kappa$ is large the result gets slightly worse, because now we cannot avoid positive weights across groups. But even if it is slightly worse, our method is still competitive.

\begin{table}[!h]
\small
\caption{RMSE for our approach with weight specification by $\kappa$- nearest-neighbors.}
\label{kappa}
\vskip -0.15in
\begin{center}
\begin{small}
\begin{tabular}{lccccc}
\hline
$\kappa$ & 2 & 3 & 4 & 5 & 6 \\
\hline
RMSE &  4.847 & 4.836 & \bf{4.828} & 4.896 &4.928 \\
\hline
\end{tabular}
\end{small}
\end{center}
\vskip -0.1in
\end{table}

\section{Applications to Predictive Modeling and GWAS of Plant Varieties}
\begin{figure}[htbp]
\begin{center}
\centerline{\includegraphics[width=\columnwidth]{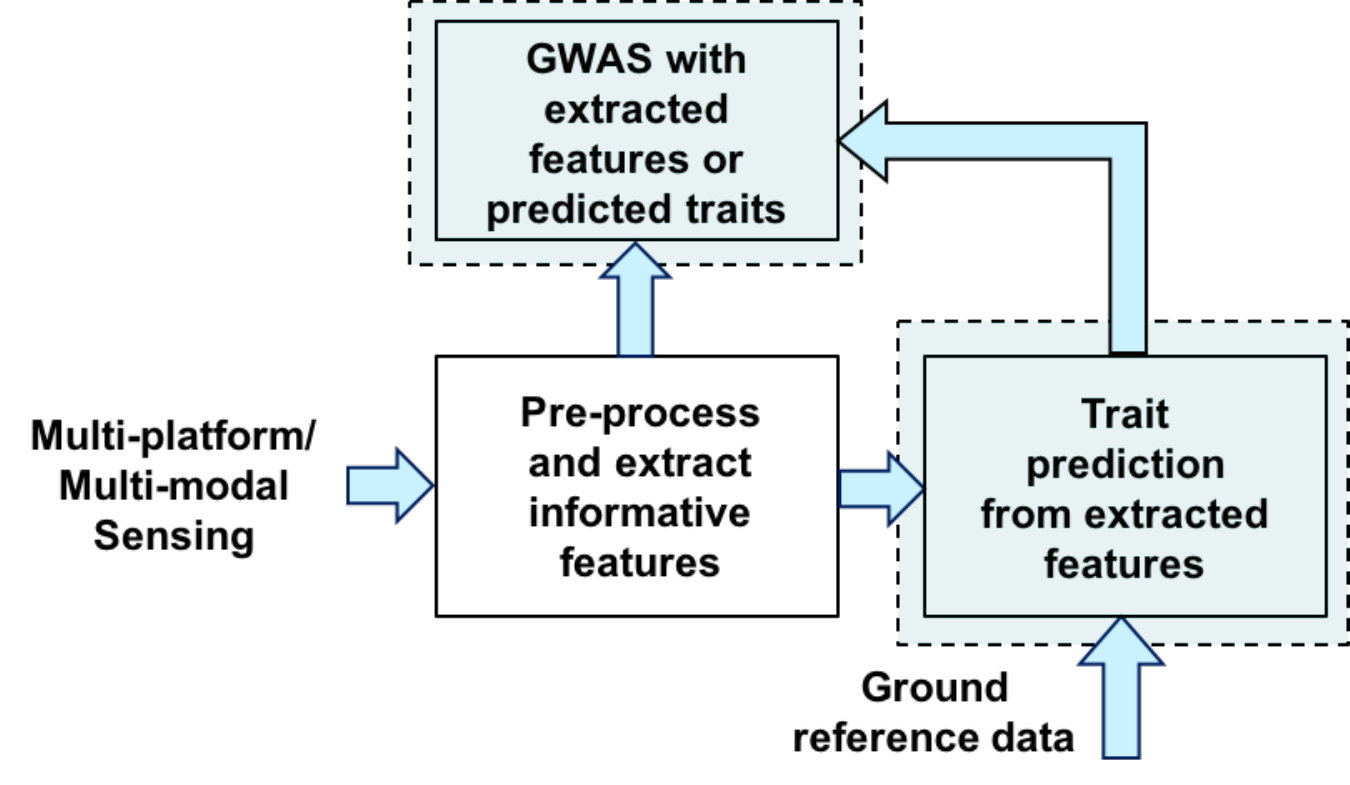}}
\caption{Context of our real data experiments. We apply our multitask learning approach for mapping from input remote sensing features to output plant traits (Section \ref{trait_prediction_expt}), and GWAS with remote sensing data features (Section \ref{GWAS_expt}).}

\label{appl_system}
\end{center}
\end{figure}

We choose high throughput phenotyping and GWAS with real data obtained from plants to demonstrate the real-world utility of our approach. Accurate phenotyping of different crop varieties is a crucial yet traditionally a time-consuming step in crop breeding requiring manual survey of hundreds of plant varieties for the traits of interest. Recently, there has been a surge in interest and effort to develop automated, high-throughput remote sensing systems \cite{tuinstra2016automated} for trait development and genome wide association studies. The typical workflow of such systems is illustrated in Figure \ref{appl_system}. Machine learning approaches are used to learn mappings between features obtained from remotely sensed images (e.g., RGB, hyperspectral, LiDAR, thermal) of plants as input and the manually collected traits (e.g., plant height, stalk diameter) as outputs. Genetic mapping is performed either with the inferred traits or directly with the features from remote sensing data.

We consider two specific problems from this pipeline: (a) predictive modeling of plant traits using features from remote sensed data (Section \ref{trait_prediction_expt}), (b) GWAS using the remote sensing data features (Section \ref{GWAS_expt}). Both these problems have unique challenges as discussed later. We apply our proposed multitask learning approach to automatically discover hierarchical groupings of output tasks, and learn differentiated models for the different groups while sharing information within groups. Apart from providing an improved predictive accuracy, the discovered groupings reveal interesting and interpretable relations between the tasks. The data used in the two experiments were collected in experimental fields 
in Summer-Fall 2015 and Summer-Fall 2016 respectively. 



\begin{figure*}[!h]
\begin{center}
\centerline{\includegraphics[width=1.6\columnwidth]{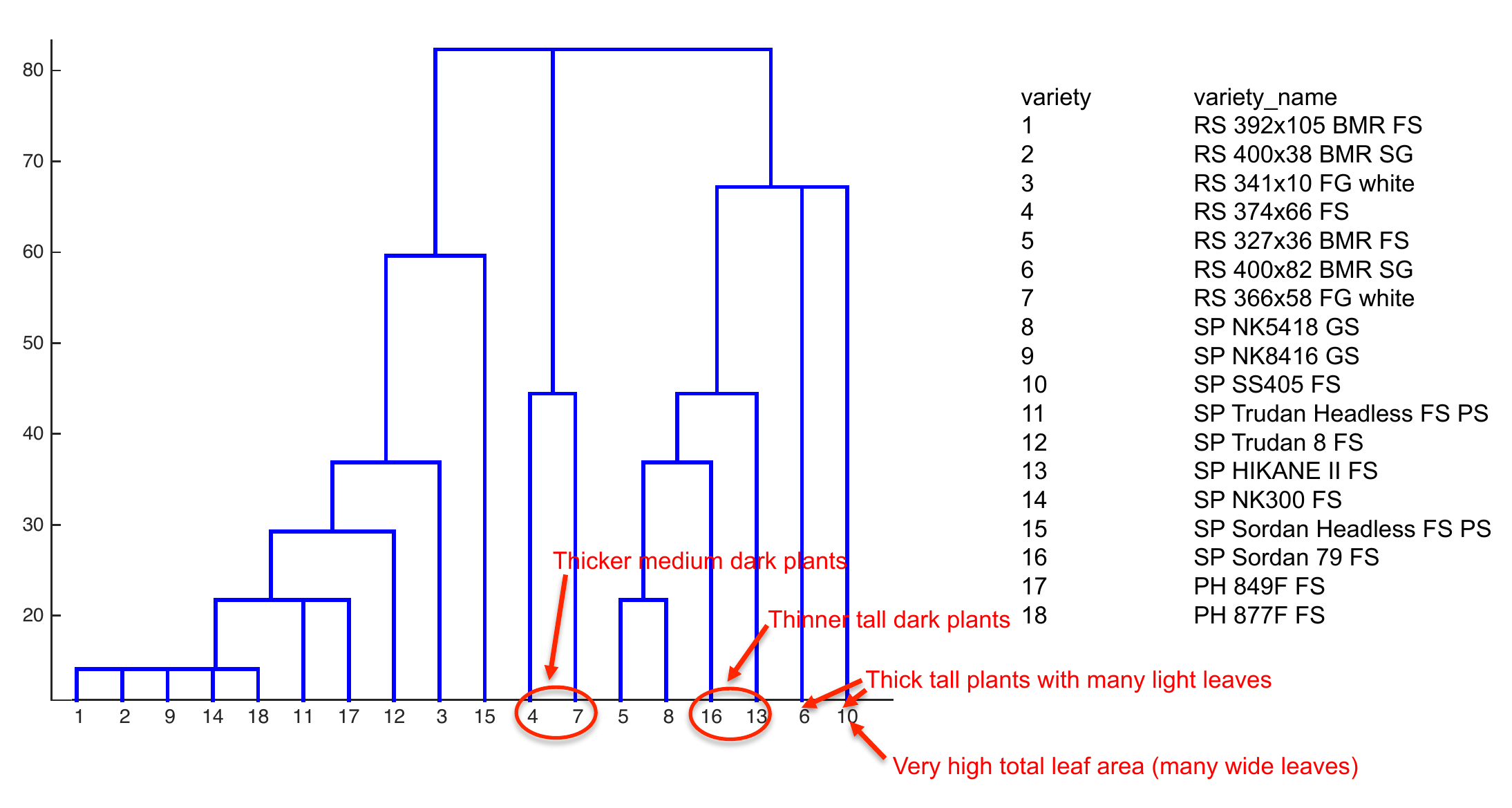}}
\caption{Tree structure of tasks (varieties) inferred using our approach for plant height.}
\label{Ph1}
\end{center}
\end{figure*}

\begin{figure*}[!h]
\minipage{0.33\textwidth}
\begin{center}
\centerline{\includegraphics[width=\columnwidth]{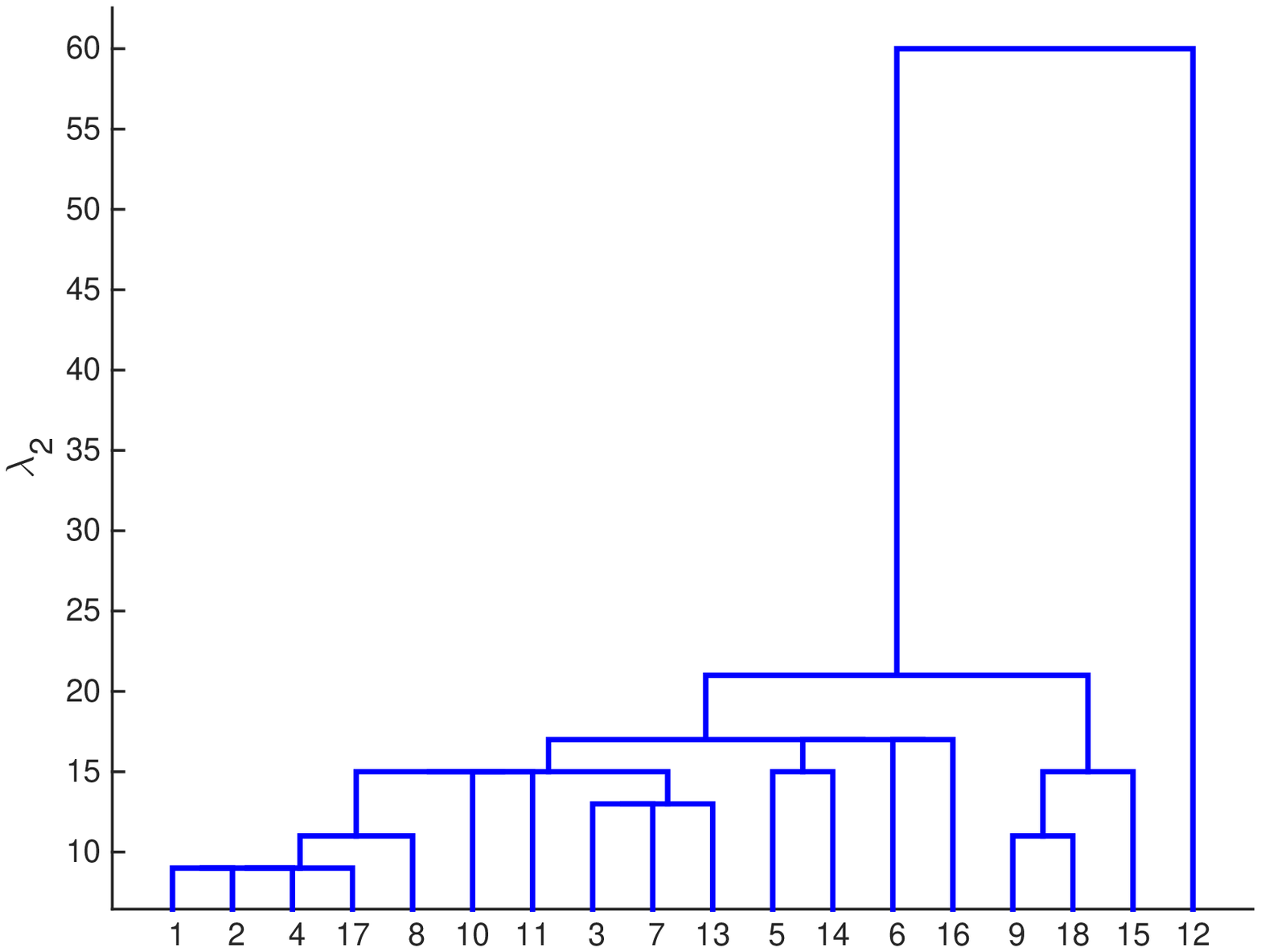}}
\caption{Tree structure of tasks (varieties) inferred using our approach for stalk diameter.}
\label{Ph2}
\end{center}
\endminipage\hfill
\minipage{0.33\textwidth}
\begin{center}
\centerline{\includegraphics[width=\columnwidth]{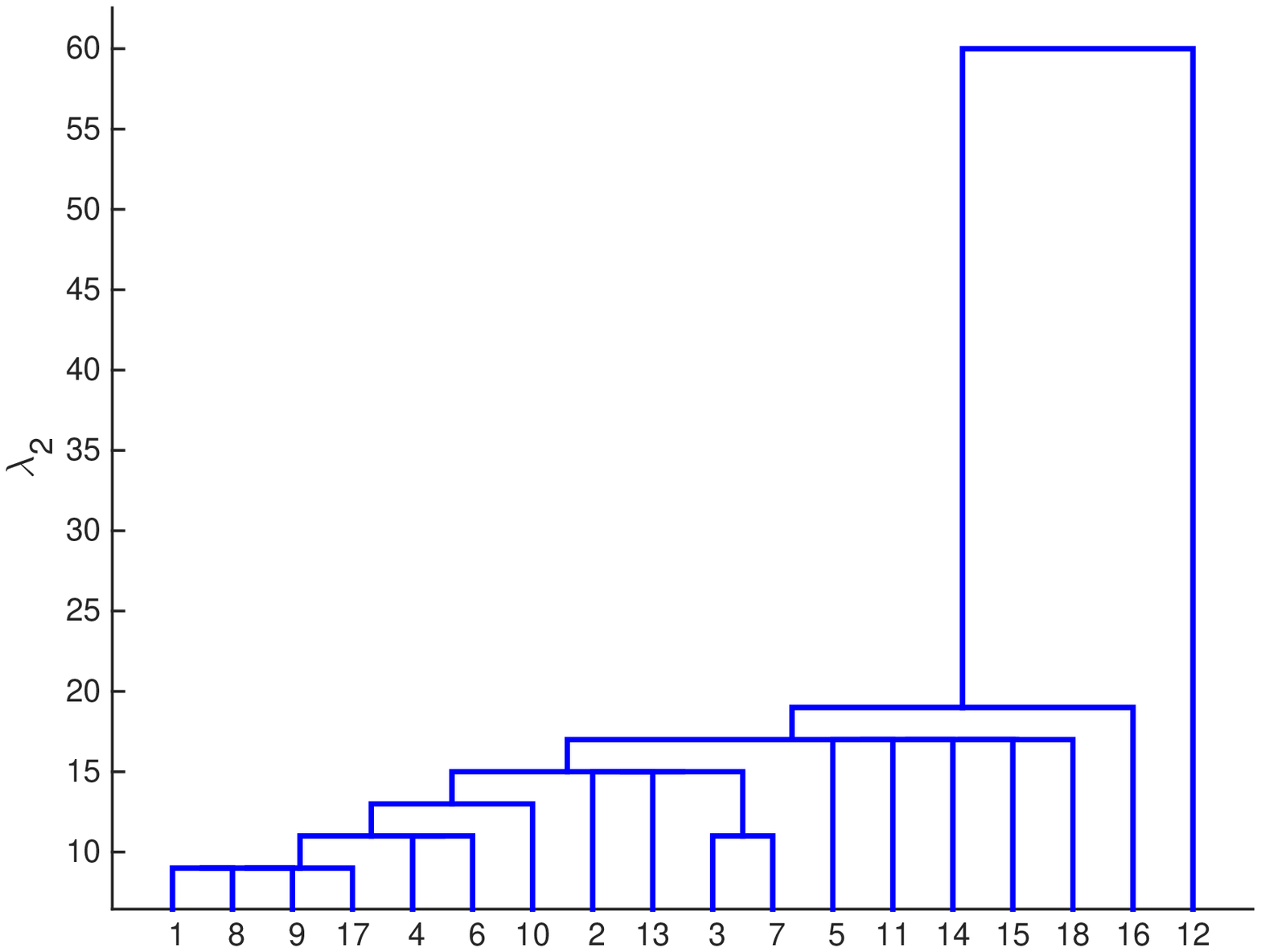}}
\caption{Tree structure of tasks (varieties) inferred using our approach for stalk volume.}
\label{Ph3}
\end{center}
\endminipage\hfill
\minipage{0.33\textwidth}
\begin{center}
\centerline{\includegraphics[width=\columnwidth]{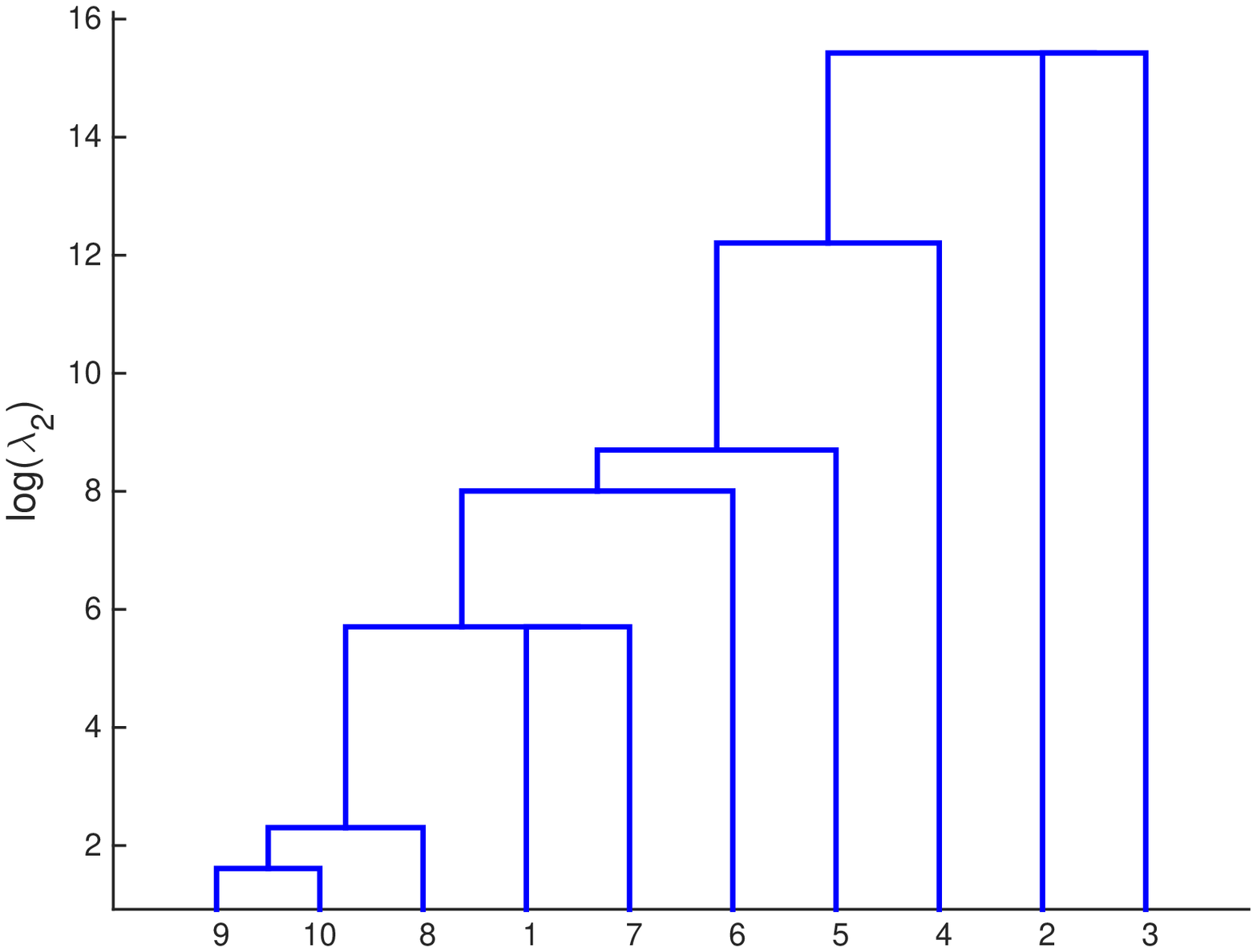}}
\caption{Tree structure of the tasks (height bins) inferred using our approach for the GWAS dataset.}
\label{GWAS}
\end{center}
\endminipage\hfill
\end{figure*}

%
\begin{table}[h]
\caption{RMSE for comparison methods on Plant Height Prediction.}
\label{RMSEphen}
\vskip 0.15in
\begin{center}
\begin{small}
\begin{tabular}{lccc}
\hline
Method & RMSE & std \\
\hline
Single model &  44.39 & 6.55 \\
No group multitask learning &  36.94 & 6.10  \\
Kang et al & 37.55 & 7.60 \\
Ours &  \bf{33.31} & 5.10 \\
\hline
\end{tabular}
\end{small}
\end{center}
\vskip -0.1in
\end{table}

\subsection{Trait Prediction from Remote Sensed Data}
\label{trait_prediction_expt}
The experimental data consists of $18$ varieties of the sorghum crop planted in $6$ replicate plot locations. From the RGB and hyperspectral images of each plot, we extract features of length $206$. Hence $n=6$, $p=206$, and $k=18$. The presence of multiple varieties with replicates much smaller in number than predictors poses a major challenge: Building separate models for each variety is unrealistic, while a single model does not fit all. This is where our hierarchical grouping and modeling approach provides the flexibility to share information that leads to learning at the requisite level of robustness. 

We have 3 different responses:  plant height, stalk diameter, and stalk volume so we get 3 tree structures. The 3 trees are given in Figure \ref{Ph1}, Figure \ref{Ph2} and Figure \ref{Ph3}. 
The cluster provide some interesting insights from a plant science perspective. As highlighted in Figure~\ref{Ph1}, for predictive models of height, thicker medium dark plants are grouped together. Similarly for thinner tall dark plants, and thick tall plants with many light leaves.
Figure \ref{Ph2} and Figure \ref{Ph3} are quite similar, which makes sense, since stalk diameter and stalk volume are highly correlated. In terms of stalk, we can see variety 12 is very different from others. It corresponds to tall thin plants with few small dark leaves.

\begin{figure*}[!h]
\minipage{0.25\textwidth}
\begin{center}
\centerline{\includegraphics[width=\columnwidth]{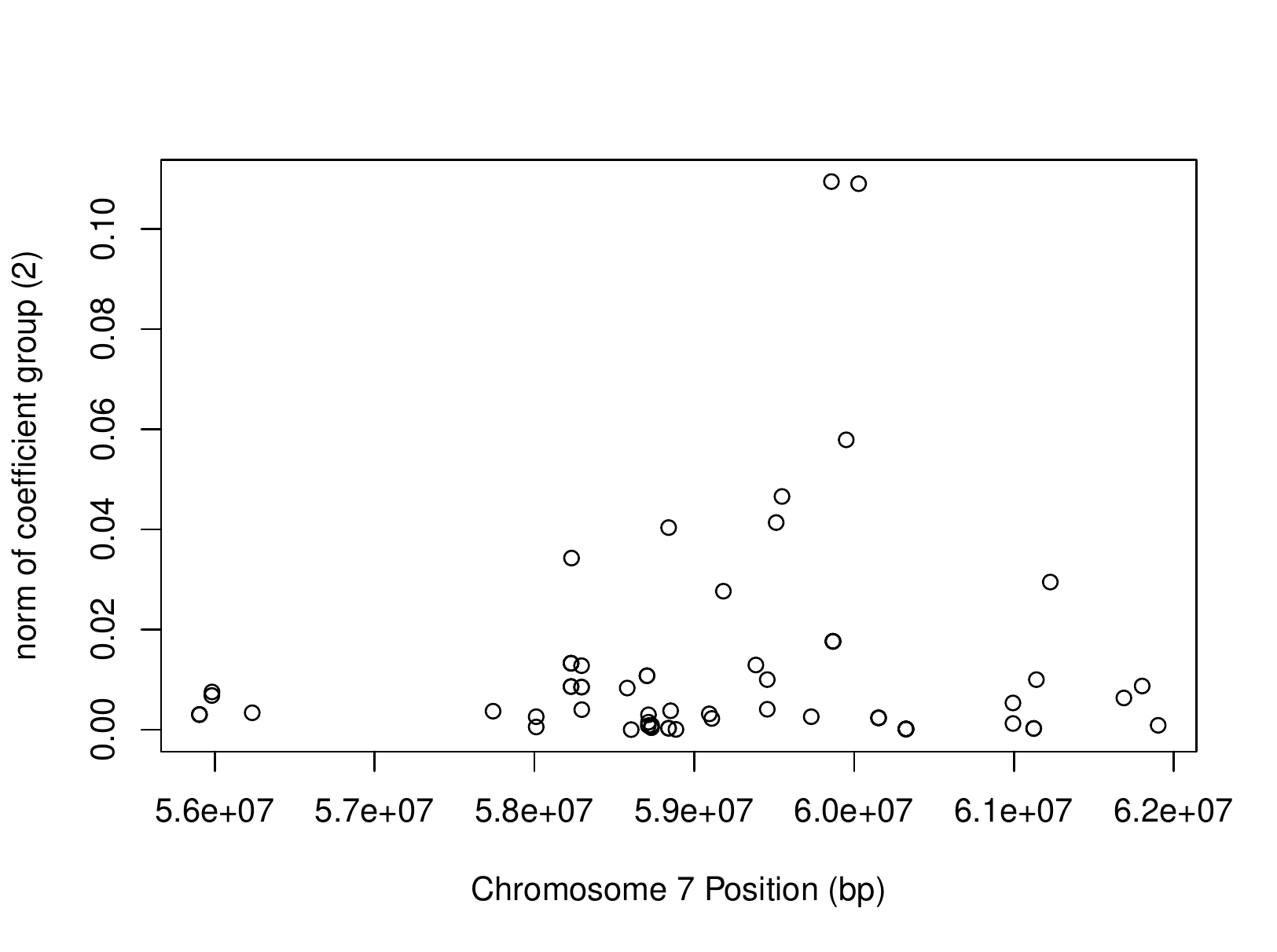}}
{Bin 2}
\label{ch72}
\end{center}
\endminipage\hfill
\minipage{0.25\textwidth}
\begin{center}
\centerline{\includegraphics[width=\columnwidth]{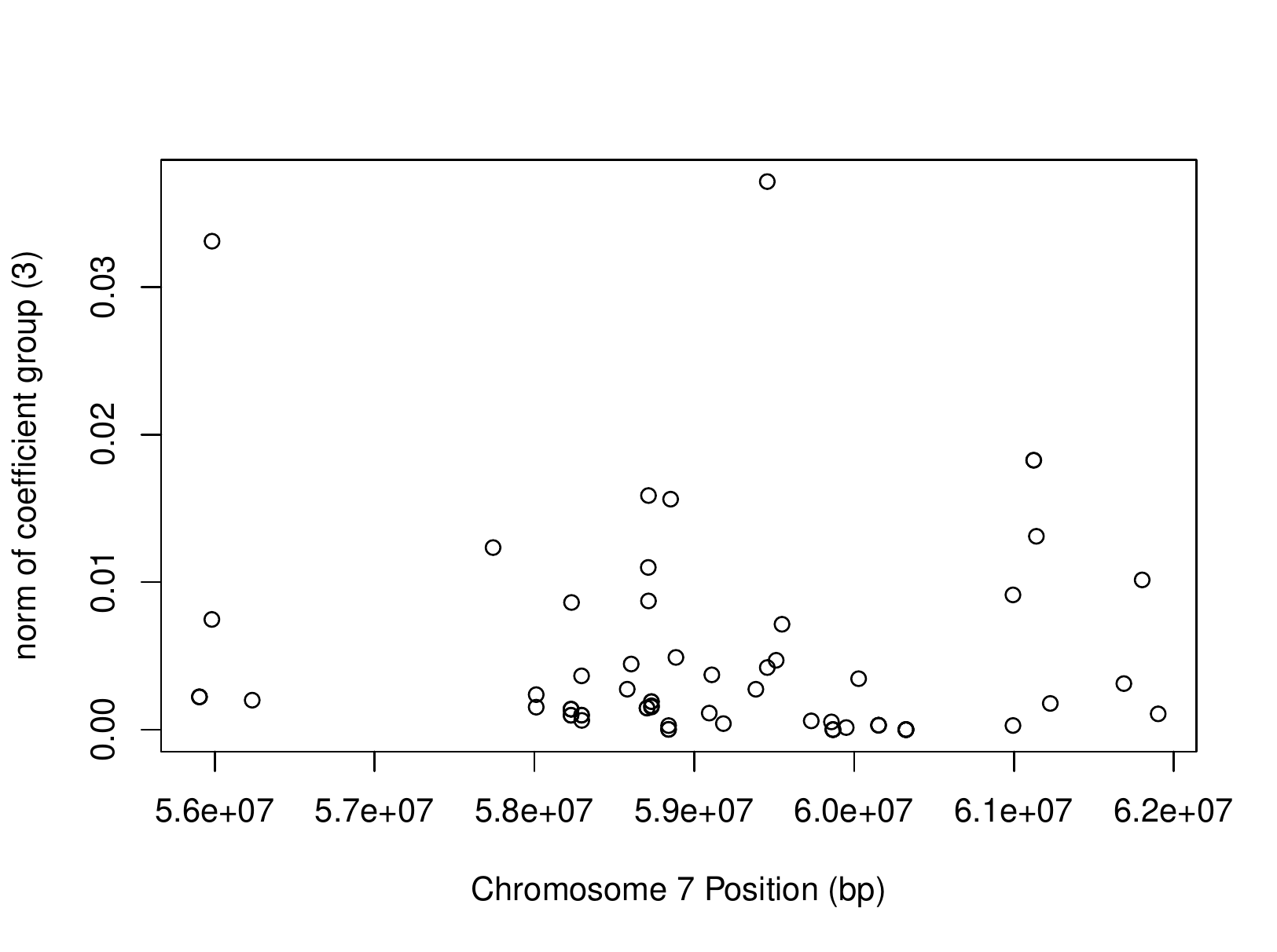}}
{Bin 3 }
\label{ch73}
\end{center}
\endminipage\hfill
\minipage{0.25\textwidth}
\begin{center}
\centerline{\includegraphics[width=\columnwidth]{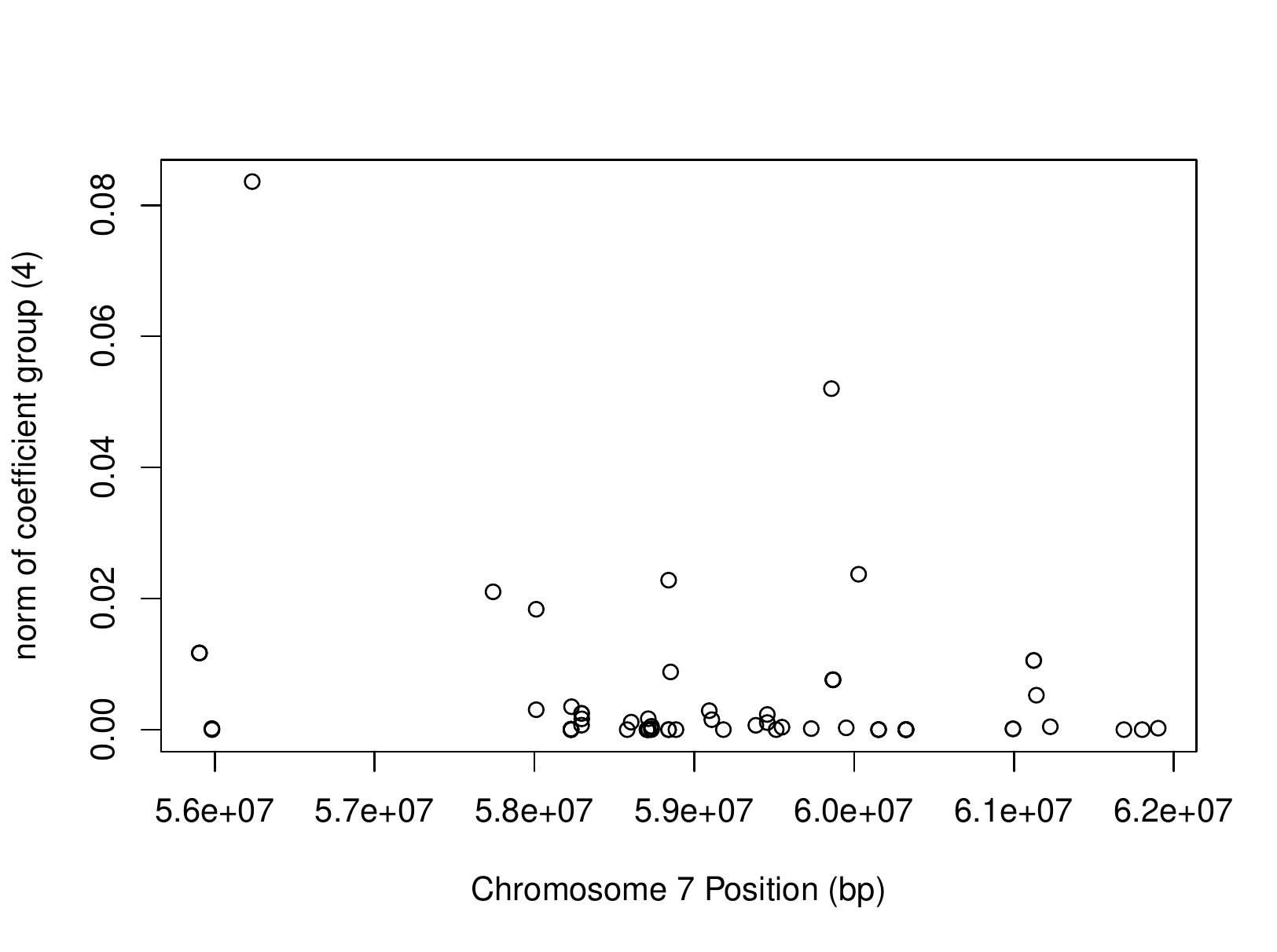}}
{Bin 4}
\label{ch73}
\end{center}
\endminipage\hfill
\minipage{0.25\textwidth}
\begin{center}
\centerline{\includegraphics[width=\columnwidth]{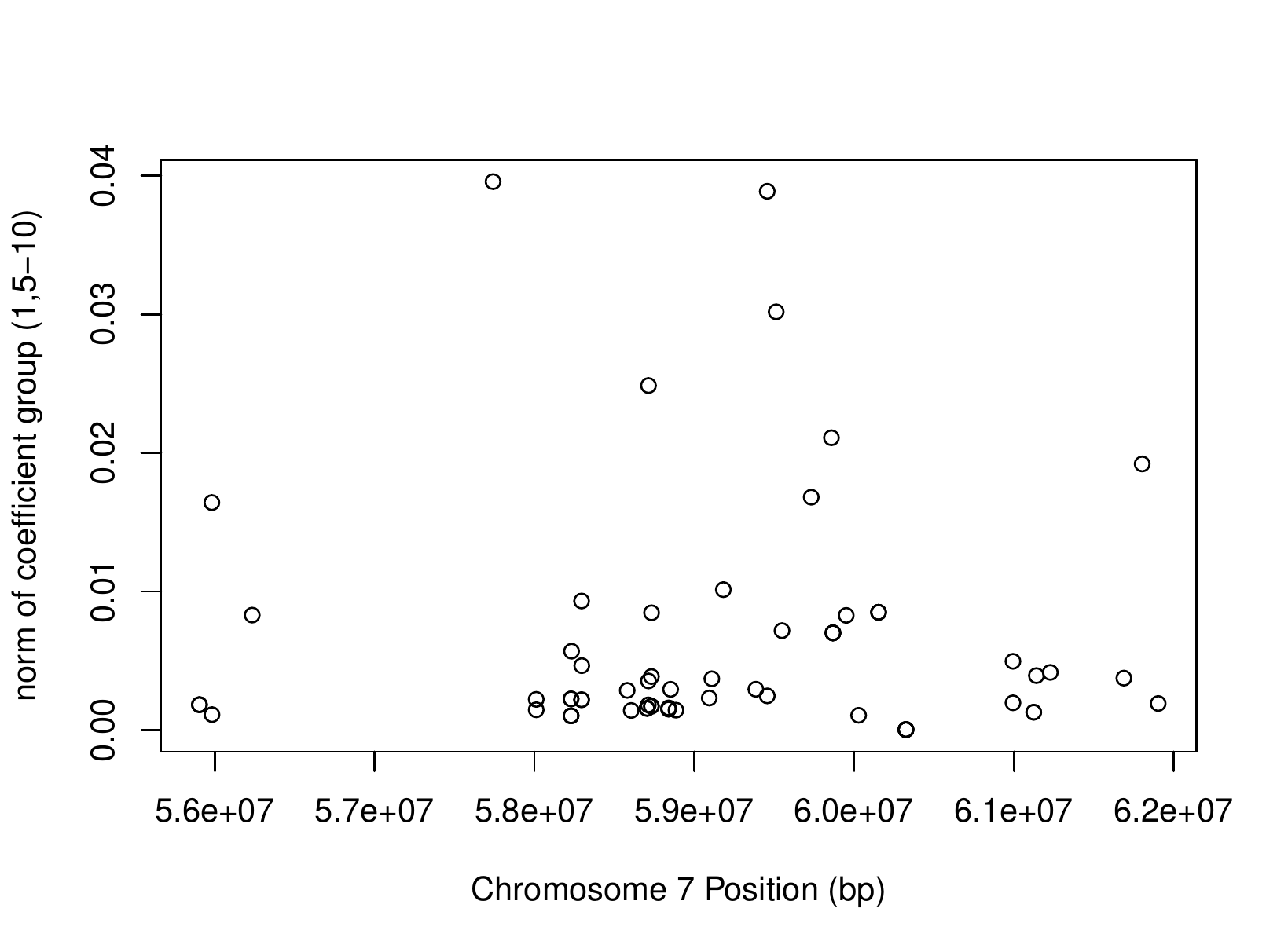}}
{Bins 1,5-10}
\label{ch7group}
\end{center}
\endminipage\hfill
\caption{Chromosome 7: Regression coefficients norm for the bin groups uncovered by our approach.  x-axis: position on the chromosome (in bp). y-axis: $\ell_2$-norm of the regression coefficients within each group.}
\end{figure*}

\begin{figure*}
\minipage{0.25\textwidth}
\begin{center}
\centerline{\includegraphics[width=\columnwidth]{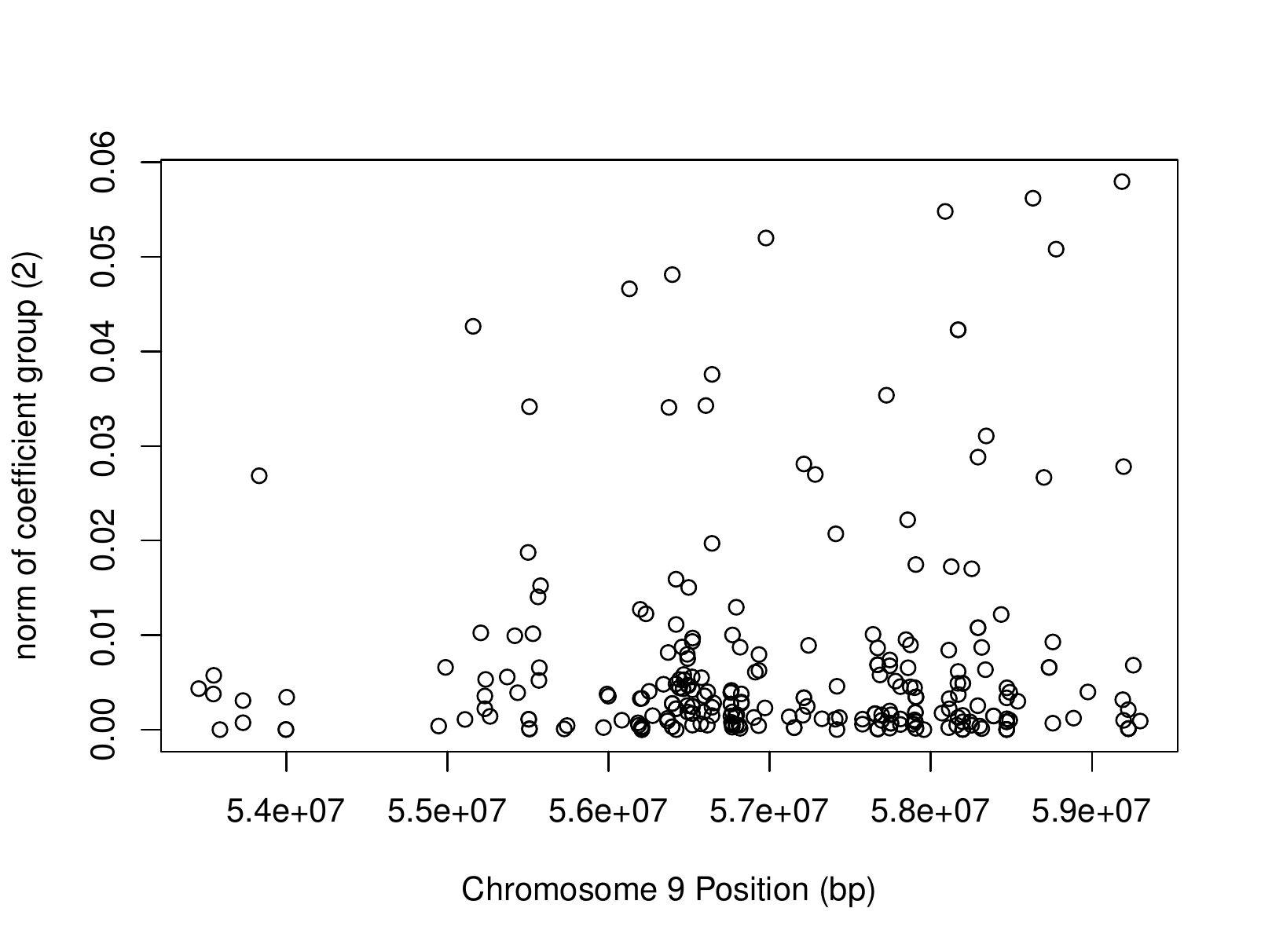}}
Bin 2
\label{ch72}
\end{center}
\endminipage\hfill
\minipage{0.25\textwidth}
\begin{center}
\centerline{\includegraphics[width=\columnwidth]{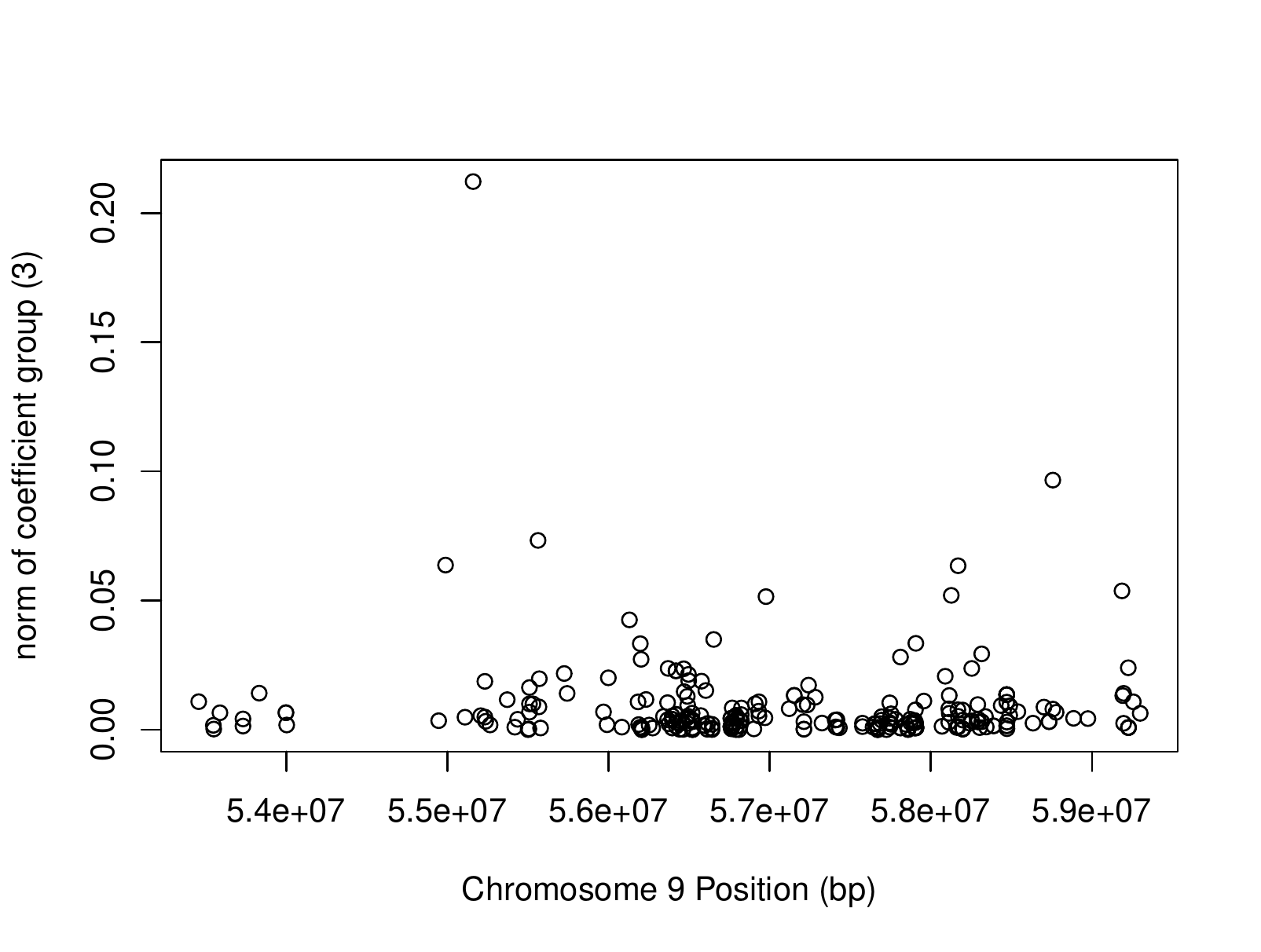}}
Bin 3
\label{ch73}
\end{center}
\endminipage\hfill
\minipage{0.25\textwidth}
\begin{center}
\centerline{\includegraphics[width=\columnwidth]{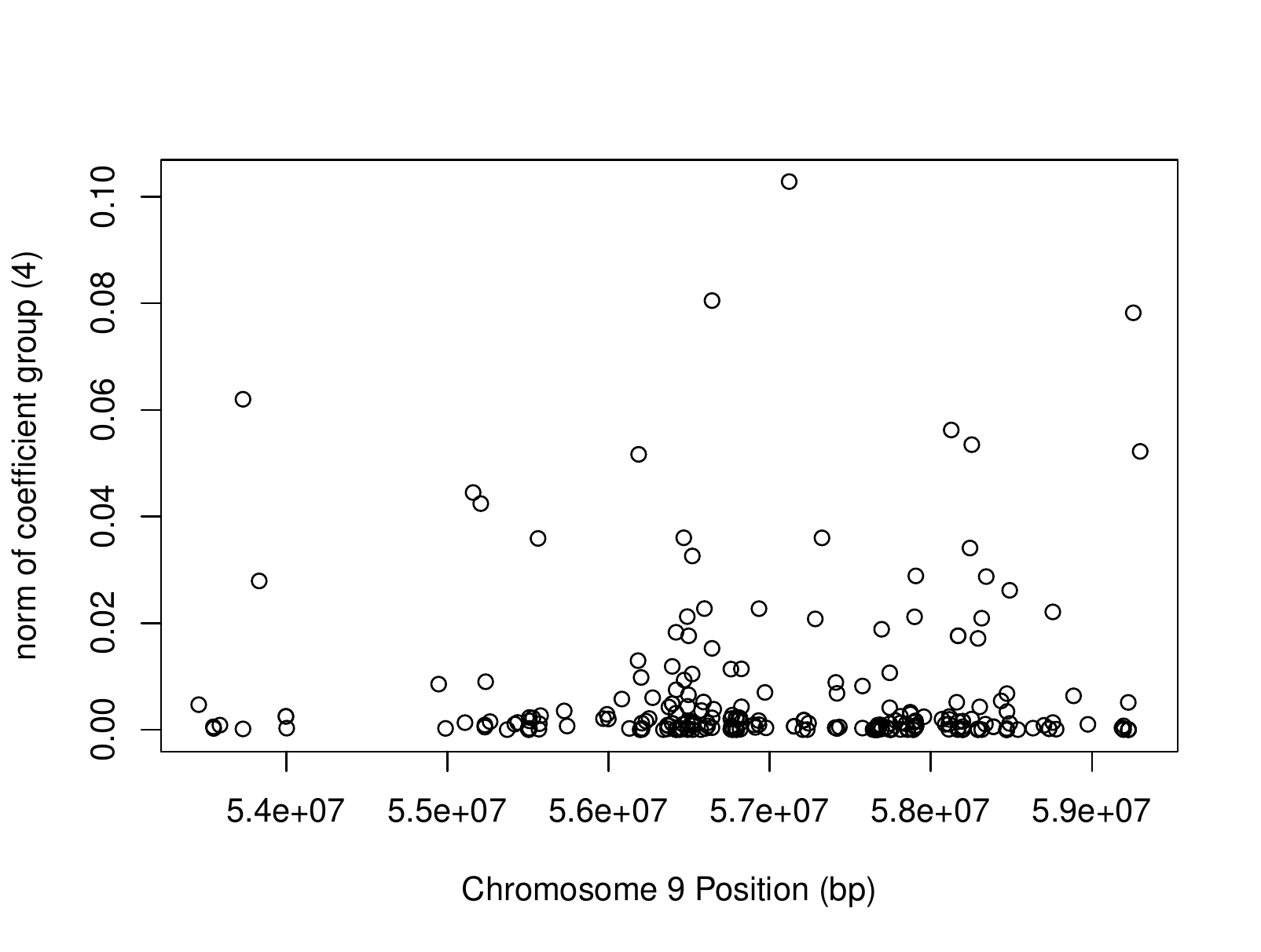}}
Bin 4
\label{ch73}
\end{center}
\endminipage\hfill
\minipage{0.25\textwidth}
\begin{center}
\centerline{\includegraphics[width=\columnwidth]{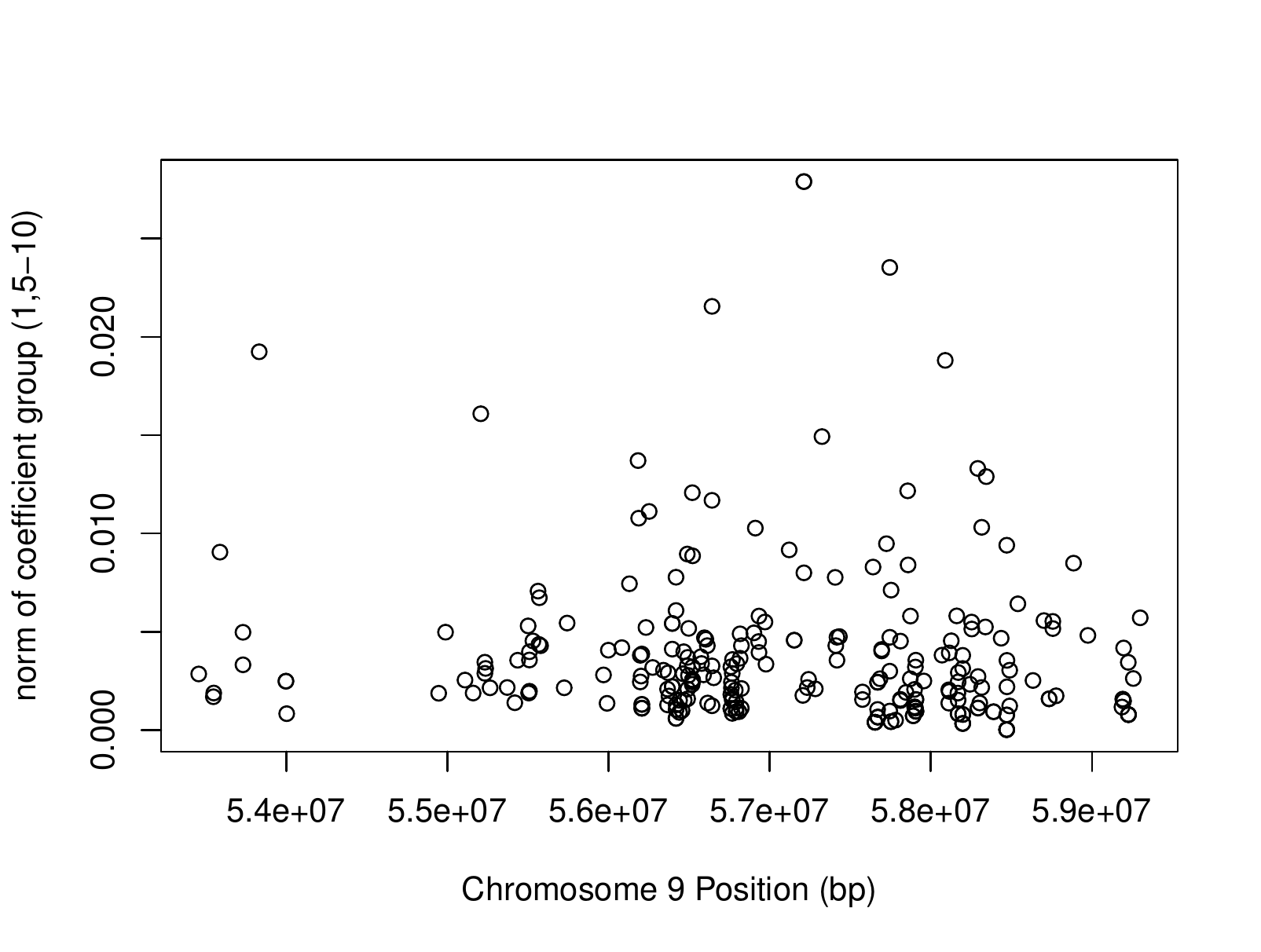}}
Bins 1,5-10
\label{ch7group}
\end{center}
\endminipage\hfill
\caption{Chromosome 9: Regression coefficients norm for the bin groups uncovered by our approach.  x-axis: position on the chromosome (in bp). y-axis: $\ell_2$-norm of the regression coefficients within each group.}
\end{figure*}

In terms of predictive accuracy comparison, we focus on predictive plant height. We perform  6-folds CV  where in each fold we make sure to include one sample from each variety. As we only have $n=6$ sample per variety (i.e. per task), it is unrealistic to learn separate models for each variety. Due to the limited sample size we  focus on assessing the quality of the groupings of various methods as follows. For each CV split, we first learn a grouping using a comparison method and then treat all the samples within a group as i.i.d and estimate their regression coefficients using Lasso.
We compare the groupings of (i)  our approach (ii) Kang et al (iii) learning a single predictive model using Lasso, treating all the varieties as i.i.d.  (iv) learning a traditional multitask model using Group Lasso (each variety form a separate group). The results are reported in Table~\ref{RMSEphen}, which indicate the superior quality of our groupings in terms of improving predictive accucacy.

\subsection{GWAS dataset}
\label{GWAS_expt}

In the second dataset, we use SNP data from 850 varieties as input for GWAS.  We considered $p=3025$ SNPs (features). There are $n=1920$ plots (observations), each containing a single variety.  The output data is the histogram of photogrammetrically derived heights obtained from RGB images of the $n=1920$ plots. We consider $10$ bins, and each bin is treated as a task. Therefore,  $k=10$. It has been demonstrated that height histograms describe the structure of the plants in the plot and are hence powerful predictors of various traits \cite{ramamurthy2016predictive}. Therefore it is worthwhile performing genomic mapping using them bypassing trait prediction. Since it is reasonable to expect the neighboring bins of the histograms to be correlated, our approach for hierarchical task grouping will result in an improved association discovery. 

For this dataset, the Kang et al. method did not scale to handle the amount of features. In general, we noticed that this algorithm is quite unstable, namely the task membership keep changing at each iteration especially as the problem dimensionality becomes large. 
The tree structure obtained by our method is given in Figure \ref{GWAS}. Please note that the $y$-axis in the figure is $\log(\lambda_2).$ We notice that bins $\{8,9,10\}$ merge quickly while bins \{2,3,4\} merge when $\lambda_2$ is extremely large. 
Note that the distance from bin 5 to bin 4 is much larger, compared to the distance from bin 7 to bin 5 (while in the figure it looks similar due to logarithmic scale. 
Bins $\{1,7,8,9,10\}$ are rarely populated. They all have small coefficients and merge together quickly; while more populated bins tend to merge later. 

We zoom in on two locations on the genome: one on chromosome 7 between positions 5.5e7 and 6.2e7 (bp) and the other on chromosome 9 between  positions 5.3e7 and 6e7 (bp). Notice that selected SNPs differ from one histogram bin group to another. Selected SNPs on chromosome 7 co-localize with ground truth (in particular with Dwarf3 gene). On chromosome  9, bin mapping identifies new regions (in addition to Dwarf1), possibly related to canopy closure, leaf distribution, flowering, and other plot-wide characteristics. As future work we plan to assess the statistical significance of the uncovered association and validate them further with domain experts.

\section{Conclusion}

In this paper we proposed a multitask learning approach that jointly learns the task parameters and a tree structure among tasks, in a fully automated, data-driven manner. The induced tree structure is natural for modeling task-relatedness and it is also easily interpretable. We developed an efficient procedure to solve the resulting convex problem and proved its numerical convergence. We also explored the statistical properties of our estimator. Using synthetic data, we demonstrated the superiority of our approach over comparable methods in terms of RMSE as well as inference of underlying task groups. Experiments with real-world high throughput phenotyping data illustrated the practical utility of our approach in an impactful application. In the future, we will explore other structured sparsity penalties, generalize the loss function to be other than squared $\ell_2,$ and pursue the applications of our method to high-throughput phenotyping further.

\appendix
\section{Proof of Proposition~\ref{prop:conv}}

We need to check that the conditions in Theorem 3.4 in \cite{combettes2008proximal} are satisfied in our case:
\begin{flalign*}
(\text{i}) &  \lim_{\|\Theta\| \to +\infty} f_1(\Theta)+ f_2(\Theta) +\sum_{s,t} f_{st}(\Theta) = +\infty  &&\\
(\text{ii}) &  (0,\ldots,0) \in \text{sri}\{(\Theta-\Theta_{(1)},\Theta-\Theta_{(2)},\Theta-\Theta_{(1)}\ldots,x-x_m) | \\
    & \qquad \qquad \qquad \Theta \in \mathbb R^{pk}, \Theta_{(1)} \in \text{dom}f_1,  \Theta_{(2)} \in \text{dom}f_2,\Theta_{(ij)} \in \text{dom}f_{ij}\} &&\\
\end{flalign*}

Let  $\mathcal H$ be the domain of $\Theta$ which can be set as $\mathbb R^{pk}$. Let $C$ be a nonempty convex subset of $\mathcal H$, the strong relative interior of $C$ is 
$$
\text{sri}(C) =  \big\{\Theta \in C | \text{cone}(C-\Theta) = \overline{\text{span}}(C-\Theta)\big\}
$$
where $\text{cone}(C) = \bigcup_{\lambda > 0}\{\lambda \Theta | \Theta \in C\}$, and $\overline{\text{span}}(C)$ is the closure of span $C$.

Now we check the conditions. For (i), $\|\Theta\|$ goes to infinity means $\|\Theta_s\|$ goes to infinity, and then we know $f_2$ goes to infinity. Therefore (i) holds.

For (ii), we do not have any restriction on $\Theta$, so the right hand side is just $\text{sri}(\mathbb R^{pk})$, hence (ii) holds.


Therefore, the proposition follows according to Theorem 3.4 of \cite{combettes2008proximal}.

\begin{acks}
The authors acknowledge the contributions of the Purdue and IBM project teams for field work, data collection/processing and discussions.

The information, data, or work presented herein was funded in part by the Advanced Research Projects Agency-Energy (ARPA-E), U.S. Department of Energy, under the Award Number DE-AR0000593. The views and opinions of authors expressed herein do not necessarily state or reflect those of the United States Government or any agency thereof.

\end{acks}

\bibliographystyle{ACM-Reference-Format}
\bibliography{paper,nrk}

\end{document}